\documentclass[runningheads]{llncs}
\usepackage[T1]{fontenc}
\usepackage{graphicx}
\usepackage{amsmath}
\usepackage{amssymb}
\usepackage[labelfont=bf]{caption}
\usepackage{cleveref}
\usepackage{subcaption}
\usepackage{tikz}
\usetikzlibrary{positioning, arrows.meta, shapes.geometric, backgrounds, fit}

\newcommand{\imagebox}[1]{%
    \makebox[1cm][c]{\includegraphics[width=1cm, height=1cm]{#1}}%
}
\usepackage{ntheorem}
\newtheorem{hyp}{Hypothesis}
\usepackage{enumitem}
\newlist{steps}{enumerate}{1}
\setlist[steps, 1]{label = Step \arabic*:}
\begin{document}
\title{Do Edges Matter? Investigating Edge-Enhanced Pre-Training for Medical Image Segmentation}
\titlerunning{Edge-Enhanced Pre-Training for Medical Image Segmentation}
%
\author{Paul Zaha\inst{1} \and
Simeon Allmendinger\inst{1,2}\orcidID{0009-0005-8741-7734} \and
Lars Böcking\inst{1,2}\orcidID{0009-0009-1365-7224} \and
Leopold Müller\inst{1,2}\orcidID{0009-0008-9968-8103} \and
Niklas Kühl\inst{1,2}\orcidID{0000-0001-6750-0876}
}
\authorrunning{P. Zaha et al.}
%
\institute{University of Bayreuth, 95447 Bayreuth , Germany \and
Fraunhofer FIT, 95447 Bayreuth, Germany}
\maketitle              

\begin{abstract}
Medical image segmentation is crucial for disease diagnosis and treatment planning, yet developing robust segmentation models often requires substantial computational resources and large datasets. 
Existing research shows that foundation models, trained on broad sets of image data and subsequently fine-tuned for specific medical tasks, can boost segmentation performance. 
However, questions remain about how particular image preprocessing steps may influence segmentation performance across different medical imaging modalities. 
In particular, edges---abrupt transitions in pixel intensity---are widely acknowledged as vital cues for object boundaries but have not been systematically examined in the pre-training of foundation models. 
We address this gap by investigating to which extend pre-training with data processed using computationally efficient edge enhancement kernels, such as kirsch, can improve cross-modality segmentation capabilities of a foundation model. 
Two versions of a foundation model are first trained on either raw or edge-enhanced data across multiple medical imaging modalities, then fine-tuned on selected raw subsets tailored to specific medical modalities. 
After systematic investigation using the medical domains Dermoscopy, Fundus, Mammography, Microscopy, OCT, US, and XRay, we discover both increased and reduced segmentation performance across modalities using edge-focused pre-training, indicating the need for a selective application of this approach. To guide such selective applications, we propose a meta-learning strategy. It uses standard deviation and image entropy of the raw image to choose between a model pre-trained on edge-enhanced or on raw data for optimal performance. Our experiments show that integrating this meta-learning layer yields an overall segmentation performance improvement across diverse medical imaging tasks by 16.42\% compared to models pre-trained on edge-enhanced data only and 19.30\% compared to models pre-trained on raw data only.

\end{abstract}

\section{Introduction}
Medical image segmentation is a cornerstone of modern diagnostics and treatment planning, as it delineates the boundaries of anatomical structures and pathological abnormalities crucial for clinical decisions. In many applications, segmenting regions of interest with high accuracy is essential for enabling effective workflows in radiology, surgery, and oncology~\cite{rajan2023gauss,muller2024redefining}. Consequently, the performance of segmentation algorithms has a direct impact on patient outcomes and healthcare efficiency~\cite{wang2022medical}. 
Training segmentation algorithms from scratch though remains computationally demanding, especially when attempting to cover the wide variety of medical imaging modalities in clinical practice~\cite{dosovitskiy2020image}. 
In response, transfer learning has become a mainstay in research, where foundation models---initially trained on large, diverse datasets---are adapted or fine-tuned to specialized tasks~\cite{Lu2015,weiss2016survey}. Commonly, these foundation models use natural-image datasets such as ImageNet for pre-training~\cite{zhao2024comparison}, and have demonstrated robust performance gains in multiple medical imaging domains~\cite{gulshan2016development,kim2018artificial,krogue2020automatic,urakawa2019detecting}. Recently, researchers have begun to explore training multiple specialized versions of these foundation models to match the distribution of different input data at inference time~\cite{pfefferle2024daft}. 
However, questions remain about how particular specialized models influence segmentation performance across different medical imaging modalities.
One fundamental technique in image analysis is edge detection, which identifies abrupt changes in pixel intensity and thereby helps define clear boundaries for objects~\cite{waghule2014overview}.
In medical contexts, these edges often correspond to structures such as bones, tumors, or vascular formations.
As such, preserving and accurately modeling edge information can be critical in developing segmentation models that generalize well to different medical conditions and imaging modalities. 
Yet, it remains unclear whether edge structures are the determining cue for segmentation models to detect abnormalities. Further, it is unkown how enhancing edge structures before or during pre-training might influence a model's ability to segment across diverse modalities. Moreover, traditional perspectives have regarded any reduction of non-edge information as disadvantageous, overlooking the possibility that removing redundant features might actually improve learning efficiency~\cite{jakubik2024data}. 
To address these concerns, we investigate whether pre-training on edge-enhanced data can systematically improve segmentation outcomes across a variety of medical imaging modalities. Specifically, we focus on the research question: To what extent does pre-training on edge-enhanced data improve segmentation quality in medical foundation models across multiple imaging modalities? For a detailed examination, we specifically focus on the usage of either raw or edge-enhanced data in pre-training instead of leveraging a mixed approach, simulating the idea of having specialized foundation models.
After systematic investigation, we discover both increased and reduced segmentation performance across modalities using edge-enhanced pre-training, indicating the need for a selective application of this approach. Consequently, we propose that meta-features of an image---namely standard deviation in pixel intensity and overall entropy---may serve as indicators of whether edge detection is beneficial. 
In the methodology, we conduct a two-stage training procedure to test hypotheses based on these meta-features. 
First, we build two versions of a general-purpose foundation model---one pre-trained on raw images and another on edge-enhanced images covering multiple medical modalities. We then fine-tune both versions on modalities. That is, specific subsets of raw data, enabling a direct comparison of their segmentation performance. 
Implementation details are kept constant to isolate the effect of edge enhancement on different types of medical images. 
Our experiments reveal that edge-focused pre-training substantially improves segmentation accuracy for some modalities but can lead to performance declines for others. Crucially, analyzing the proposed meta-features helps to identify which images are likely to benefit from edge-focused pre-training versus those that perform better with raw-data pre-training. By leveraging this information, we develop a meta-feature classifier that selects which foundation model should be applied to maximize segmentation quality.

\section{Methodology}
\begin{figure}[ht]
\centering

\begin{subfigure}[t]{0.95\textwidth}
\centering
\resizebox{\textwidth}{!}{
\begin{tikzpicture}[
    node distance=1 cm and 1 cm,
    box/.style={draw, rounded corners, minimum height=1cm, minimum width=3 cm, align=center, fill=white},
    arrow/.style={-{Stealth}, thick},
    dashedarrow/.style={-{Stealth}, thick, dashed, gray!70},
    label/.style={font=\footnotesize},
    backgroundBox/.style={ rounded corners, draw=black, inner sep=0.3cm, dashed},
    backgroundBoxOuter/.style={fill=gray!10, rounded corners, draw=black!30, inner sep=0.4cm, inner ysep=1cm}
]

\node[box] (rawdata) {
    Raw Data \\[0.3cm]
    $\mathcal{D} = \{(x_i, y_i)\}_{i=1}^N$ \\[0.3cm]
    \begin{minipage}{3.5cm}
        \centering
        \includegraphics[width=1cm]{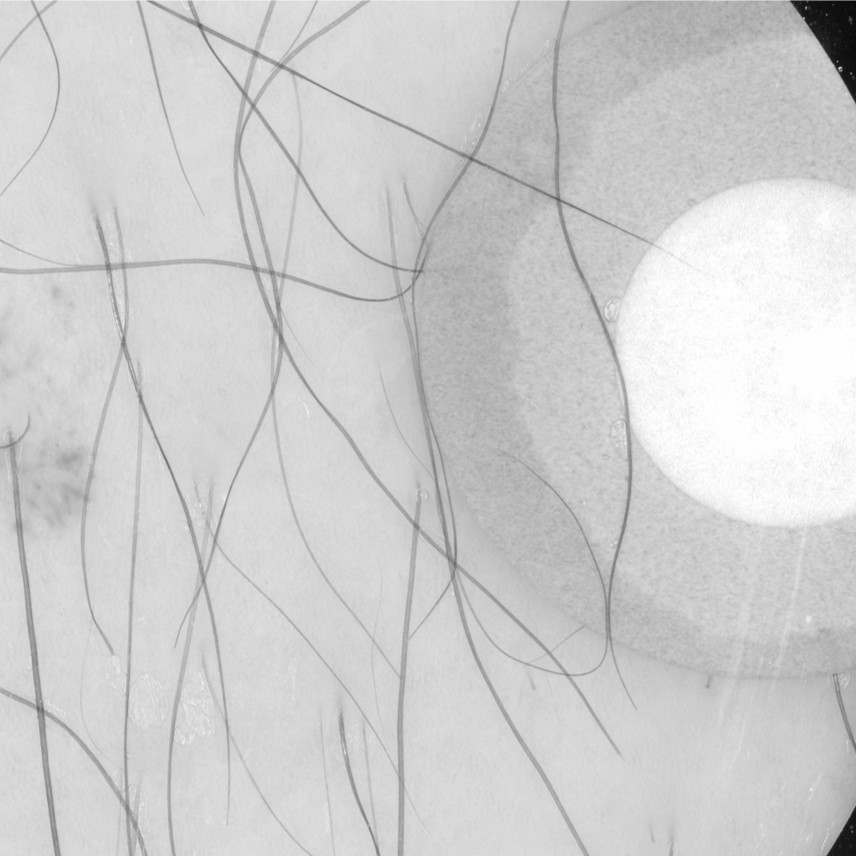}
        \includegraphics[width=1cm]{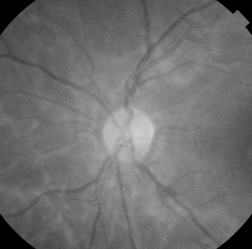}
        \includegraphics[width=1cm]{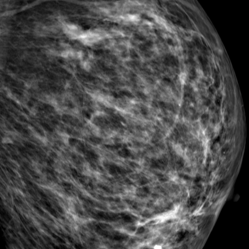}\\
        \includegraphics[width=1cm]{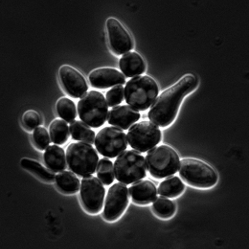}
        \includegraphics[width=1cm]{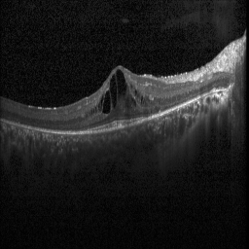}
        \includegraphics[width=1cm]{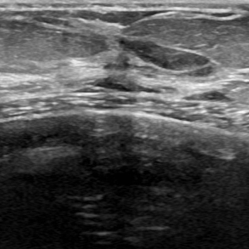}\\
        \includegraphics[width=1cm]{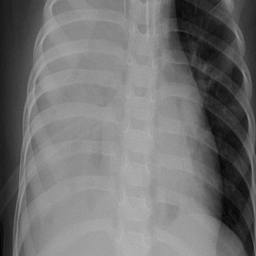}
    \end{minipage}
};
\node[box, right =1 cm of rawdata] (edgedata) {
    Edge-enhanced \\ $E(x_i)$\\[0.3cm]
    \begin{minipage}{3.5cm}
        \centering
        \imagebox{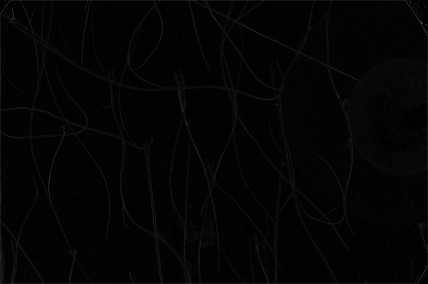}
        \imagebox{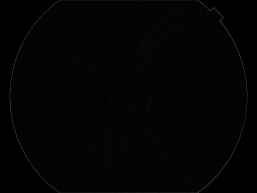}
        \imagebox{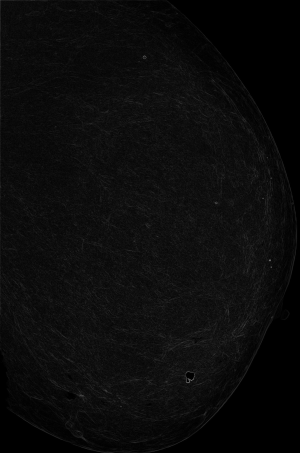}\\
        \imagebox{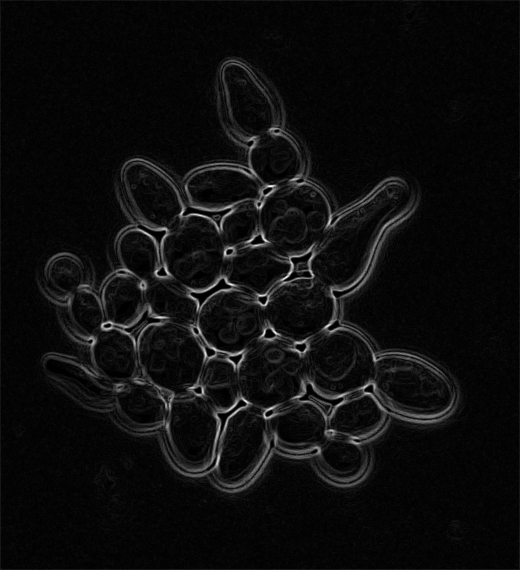}
        \imagebox{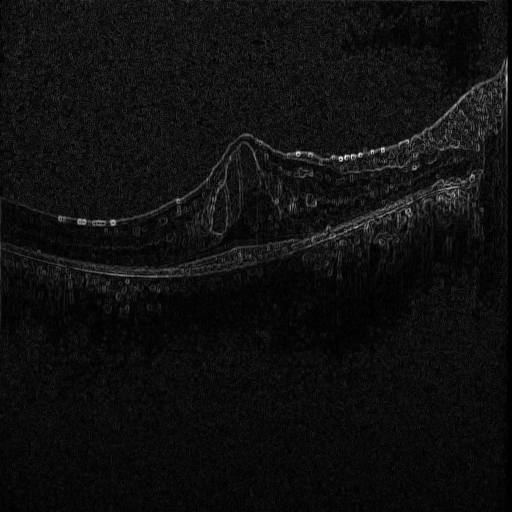}
        \imagebox{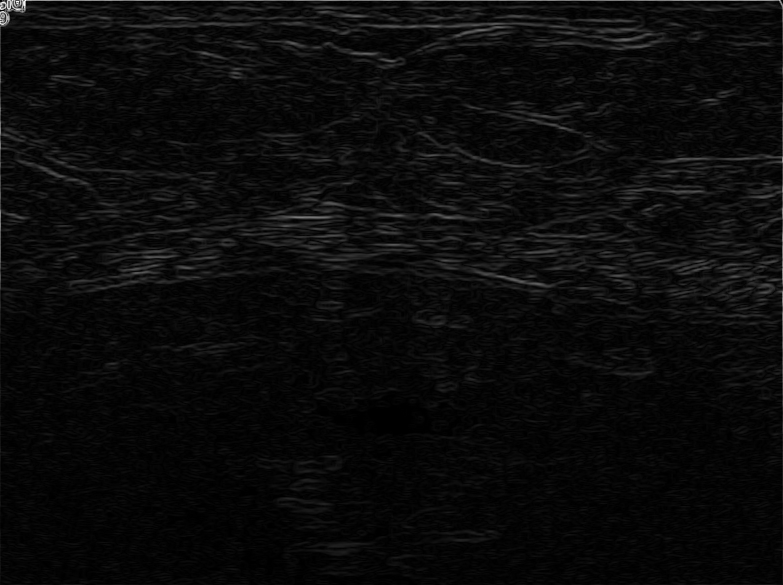}\\
        \imagebox{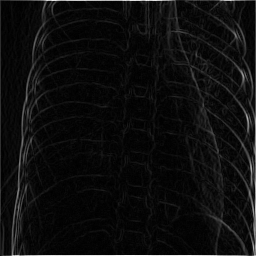}
    \end{minipage}
};

\node[box, below=1cm of edgedata] (meta_feat) {Extract $\phi(x_i)$};

\node[box, above right=0cm and 1cm of edgedata] (train_raw) {Train $f_\theta$ \\ on $\{(x_i, y_i)\}^N_{i=1}$};
\node[box, right=1cm of edgedata] (train_edge) {Train $f^\star_\theta$ \\ on $\{(E(x_i), y_i)\}^N_{i=1}$};

\node[coordinate, above=1.5cm of rawdata] (boxpad) {}; 
\node[box, right=2cm of train_raw] (finetune_raw) {Fine-tune $f_\theta$ on \\ modality data $\mathcal{D}^{mod}$};
\node[box, right=2cm of train_edge] (finetune_edge) {Fine-tune $f^\star_\theta$ on \\ modality data $\mathcal{D}^{mod}$};

\node[box, below right=1cm and 2cm of finetune_raw] (compare_perf) {Compare perf. \\ $\delta_i = \mathbb{I}[f^\star_{\theta} > f_{\theta}]$};
\node[box, below=3.3cm of compare_perf] (train_meta) {Map meta-classifier \\ $m(\phi(x_i)) \rightarrow{} \delta_i$};

\draw[arrow] (rawdata.south) |- (meta_feat);
\draw[arrow] (rawdata) -- (edgedata);
\draw[arrow] (edgedata) -- (train_edge);
\draw[arrow] (train_edge) -- (finetune_edge);
\draw[arrow] (rawdata.north) |- (train_raw);
\draw[arrow] (train_raw) -- (finetune_raw);
\draw[arrow] (finetune_raw) |- (compare_perf);
\draw[arrow] (finetune_edge) |- (compare_perf);
\draw[arrow] (meta_feat) -- (train_meta);
\draw[arrow] (compare_perf) -- (train_meta);

\begin{pgfonlayer}{background}
    \node[backgroundBoxOuter, fit=(rawdata) (train_meta) (boxpad)] {};
    \node[backgroundBox, fit=(train_raw) (train_edge)] {};
    \node[backgroundBox, fit=(finetune_raw) (finetune_edge)] {};
    \node[backgroundBox, fit=(compare_perf) (train_meta)] {};
    
\end{pgfonlayer}

\node[label, above=0.5cm of train_raw] {\textbf{General pre-training}};
\node[label, above=0.5cm of finetune_raw] {\parbox{5cm}{\centering\textbf{Modality-specific fine-tuning\\for every modality}}};

\node[label, above=0.5cm of compare_perf] {\textbf{Meta-classifier training}};
\end{tikzpicture}
}
\caption{Training: Our approach to pre-training, modality-specific fine-tuning, and meta-classifier mapping.}
\label{fig:meta-model-training}
\end{subfigure}

\begin{subfigure}[t]{0.95\textwidth}
\centering
\resizebox{\textwidth}{!}{
\begin{tikzpicture}[
    node distance=1 cm and 1 cm,
    box/.style={draw, rounded corners, minimum height=1cm, minimum width=3 cm, align=center, fill=white},
    arrow/.style={-{Stealth}, thick},
    label/.style={font=\footnotesize},
    backgroundBoxOuter/.style={fill=gray!10, rounded corners, draw=black!30, inner sep=0.5cm, inner ysep=1cm}
]

\node[box] (input) {New Image \\ $x_i$};
\node[coordinate, above=1.5cm of input] (boxpad) {}; 
\node[coordinate, below=1.5cm of input] (boxpadbottom) {}; 
\node[box, right=1.5cm of input] (feat_inf) {Extract $\phi(x_i)$};
\node[box, right=1.5cm of feat_inf] (meta_clf) {Meta-classifier \\ $m(\phi(x_i))$};
\node[box, above right=0.5cm and 1.5cm of meta_clf] (infer_raw) {Apply $f_\theta$};
\node[box, below right=0.5cm and 1.5cm of meta_clf] (infer_edge) {Apply $f^\star_\theta$};
\node[box, right=3cm of meta_clf] (output) {Segmentation Output \\ 
$\hat{y_i} =
\begin{cases}
    f_{\theta}(x_i), & \text{if } m(\phi(x_i)) = 0 \\
    f^\star_{\theta}(x_i), & \text{if } m(\phi(x_i)) = 1
\end{cases}$};

\draw[arrow] (input) -- (feat_inf);
\draw[arrow] (feat_inf) -- (meta_clf);
\draw[arrow] (meta_clf) -- node[label, above left] {$0$} (infer_raw);
\draw[arrow] (meta_clf) -- node[label, below left] {$1$} (infer_edge);
\draw[arrow] (infer_raw) -- (output);
\draw[arrow] (infer_edge) -- (output);

\begin{pgfonlayer}{background}
    \node[backgroundBoxOuter, fit=(input) (output) (boxpad) (boxpadbottom)] {};
\end{pgfonlayer}

\end{tikzpicture}
}
\caption{Inference: The meta-classifier $m(\phi(x_i))$ decides whether to use $f_{\theta}$ or $f^\star_{\theta}$ at inference time.}
\label{fig:meta-model-inference}
\end{subfigure}

\caption {Our approach: (a) training pipeline and (b) inference pipeline.}
\label{fig:meta-model-pipeline}
\end{figure}
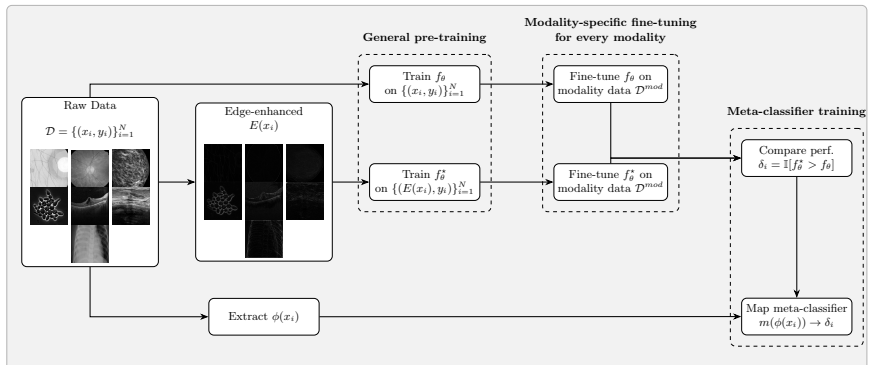
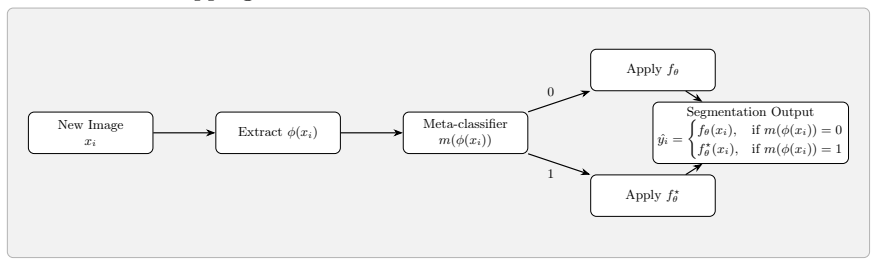

Our approach introduces a minor adaptation to the segmentation pipeline by incorporating edge-enhancement as a pre-processing step for cross-modality data in pre-training.
We define a dataset $\mathcal{D} = \{(x_i, y_i)\}_{i=1}^N$, where $x_{i}$ is an input image part of the image space $\mathcal{X}$ and $y_{i}$ is the ground truth segmentation mask part of the segmentation space $\mathcal{Y}$. Both $x_i$ and $y_i$ are parameterized by the spatial height $m$ and the width $n$, with $m,n\in\mathbb{N}^{+}$.
We introduce an edge enhancement function $E: \mathcal{X} \rightarrow \mathcal{X}_{E}$ that transforms every input image $x_{i}$ into its edge-enhanced representation $x_{i,E}$. 
The model pre-trained on raw data is denoted as $f_{\theta}: \mathcal{X} \rightarrow \mathcal{Y}$, which is parameterized by $\theta$ and maps from the image space $\mathcal{X}$ to the same segmentation space $\mathcal{Y}$. The model pre-trained on edge-enhanced data is denoted as $f_{\theta}^{\star}: \mathcal{X}_{E} \rightarrow \mathcal{Y}$, parameterized by $\theta$ and maps from the edge-enhanced image space $\mathcal{X}_{E}$ to the segmentation space $\mathcal{Y}$.
The models are first pre-trained on diverse medical domains and then fine-tuned separately for individual domains represented by subsets of data $\mathcal{D}^{mod}\in\mathcal{D}$ for every modality. 
The segmentation quality is evaluated using a performance metric $\mathcal{P}$, which quantifies the discrepancy between the predicted segmentation mask $\hat{y_i}$ and the ground truth $y_{i}$. 
Our hypothesis is formalized as: \[\mathbb{E}_{(E(x_i), y_i) \sim \mathcal{D}} \left[ \mathcal{P}(f_{\theta}^{\star})\right] > \mathbb{E}_{(x_i, y_i) \sim \mathcal{D}} \left[ \mathcal{P}(f_{\theta}) \right],
\] where $\mathbb{E}$ denotes the expectation over the the dataset $\mathcal{D}$.
Our interest lies in the pre-training process, where we replace raw inputs $(x_i,y_i)$ with edge-enhanced data $(E(x_i),y_i)$. 
For edge enhancement, we employ the Kirsch filter~\cite{kirsch1971computer}, which applies eight convolution kernels to detect edges in different orientations. As disclosed in~\cref{fig:meta-model-training}, the pre-training is followed up with fine-tuning the foundation model on domain-specific subsets $\mathcal{X}^{mod} \subset \mathcal{X}$, where each subset $\mathcal{X}^{mod}$ corresponds to a specific medical domain. This results in two fine-tuned models for each domain: one pre-trained on raw images $\mathcal{X}$ and the other on edge-enhanced images $\mathcal{X}_{E}$. To evaluate the segmentation performance $\mathcal{P}$ across different medical domains, we report the average of the Dice Similarity Coefficient (DSC) and the Normalized Surface Distance (NSD). Determining whether to use the model pre-trained on raw images $\mathcal{X}$ or the one pre-trained on edge-enhanced images $\mathcal{X}_{E}$ for achieving the best results for a given input is non-trivial. As our results will demonstrate, there is no universal solution to this problem, since models rarely exhibit increased segmentation performance over others for all input images. To address this, we not only analyze differences in segmentation quality but also introduce a meta-learning strategy. This strategy leverages meta-features extracted from the image data to predict when edge-enhanced pre-training is beneficial for foundation models.
We define $\phi(x_i)\in\mathbb{R}^d$ as a meta-feature vector of the image $x_i$. A meta-classifier $m:\mathbb{R}^d\rightarrow\{0,1\}$ learns a mapping to approximate $\delta_i = \mathbb{I}\left[ P(f^\star_{\theta}) > P(f_{\theta}) \right]$, choosing the superior segmentation pipeline based on $\phi(x_{i})$. The approach is visualized in~\cref{fig:meta-model-training}. To create the feature vector $\phi(x_i)$, we set up the following two hypotheses to explore how various pixel intensity properties---standard deviation and entropy---affect segmentation quality when edge detection is applied as pre-training step. We select these meta-features as they capture distinct properties of image structure that are crucial for the behavior of edge operators~\cite{el2012new,jassim2013semi}. First, with edge kernels transforming abrupt pixel intensity changes into bright pixels while rendering homogenous regions darker, a broad distribution of pixel intensities suggests a diverse range of features, which could enhance the density of informative edge structures in the processed image.
\begin{hyp}
    Images with higher standard deviation in pixel intensity exhibit improved segmentation quality when the foundation model is trained on images pre-processed using edge detection kernels.
    \label{hyp:sd}
\end{hyp}
We consider the \textit{standard deviation} of the image intensity values, which measures the variability in pixel intensities across the image. Images with higher standard deviation tend to exhibit more pronounced edges and textures, which we hypothesize positively correlates with segmentation performance (H\ref{hyp:sd}). We compute the standard deviation $\sigma_{x}=\sqrt{\frac{1}{m*n} \sum^{m}_{i=1} \sum^{n}_{j=1} (x_{i,j}-\mu_{x})^{2}}$ of the image $x \in \mathbb{R}^{m \times n}$. 
Second, while information reduction through edge detection can have a substantial impact on model performance, it may be advantageous if it removes non-informative or redundant features.
However, this potential improvement is contingent upon the presence of a sufficient information density in the image. One way to quantify this density is image entropy, which measures the overall information content. We hypothesize that higher entropy images, which inherently contain more structural details, benefit more from edge-based pre-processing:
\begin{hyp}
    Images with higher entropy exhibit improved segmentation quality when the foundation model is trained on images pre-processed using edge detection kernels.
    \label{hyp:entropy}
\end{hyp}

We evaluate the \textit{image entropy} $\varepsilon_x$, which quantifies the complexity and information content of the image as $\varepsilon_{x} = -\sum_{i=1}^{m}\sum^{n}_{j=1} p(i,j) * \log_{2}p(i,j)$. Higher entropy indicates greater variability and detail in the image, which we hypothesize positively correlates with segmentation performance (H\ref{hyp:entropy}). By incorporating these features into our segmentation pipeline, we aim to provide a principled approach for selecting the most effective pre-training strategy based on the characteristics of the input data. For inference, we predict $\hat{y_i}$ with the meta-classifier $m$ determining the segmentation pipeline as presented in~\cref{fig:meta-model-inference}. We denote the full meta-model pipeline composed of $m$, $f_{\theta}$ and $f_{\theta}^{\star}$ as $f_{\theta}^{meta}$.\\
\textbf{Experiment.} The dataset $\mathcal{D}$ utilized in this study originates from the ``CVPR 2024: Segment anything in medical images on laptop'' challenge~\cite{dataset}. For validity of the evaluation framework, the dataset was initially partitioned into 80\% training data and 20\% testing data. Ensuring robustness of the approach and showing potential also on smaller datasets, a subset of the training data was chosen through stratified random sampling, achieving representative distributions across different imaging modalities. For all imaging techniques, including X-ray, ultrasound (US), mammography, optical coherence tomography (OCT), fundus, dermoscopy, and microscopy, 10\% of the training sample was used for training. The sample sizes between modalities in the testing split were aligned for improved comparability.
For the segmentation task, the EfficientVIT-SAM~\cite{zhang2024efficientvit} architecture
was chosen due to its computational efficiency and state-of-the-art performance in medical image segmentation~\cite{xu2024esp,pfefferle2024daft,stock2024segment}.
To enhance the model's feature extraction capabilities, knowledge distillation was employed before the main training phase. Specifically, pretrained representations from TinyViT~\cite{wu2022tinyvit},
as used in LiteMedSam~\cite{ma2024segment}, were transferred to the EfficientViT-SAM encoder.
Knowledge distillation was conducted for 20 epochs with a batch
size of 8 for improved generalization and avoidance of overfitting. 
Following the distillation step, this model was trained for 20 epochs using a larger batch size of 48 due to efficiency twice, once using $X$ and once using $X_{E}$, representing the pre-training step. Both pre-training and fine-tuning protocols used the unweighted sum of binary cross-entropy loss $\mathcal{L}_{BCE}$, dice loss $\mathcal{L}_{Dice}$ and intersection over union loss $\mathcal{L}_{IoU}$.
Upon completion of pre-training, the two foundation models were further fine-tuned for 20 epochs with a batch size of 48 using raw, modality-specific data $\mathcal{X}^{mod}\subset \mathcal{X}$. For hardware acceleration, an Nvidia RTX 4090 (Laptop) was used during the whole experiment. The meta-model mapping $m$ is generated on 80\% of the testing data using discrete optimization, and evaluated using the remaining 20\% of the testing data.
\section{Results}
\begin{figure}[h]
    \centering
    \includegraphics[width=0.85\linewidth]{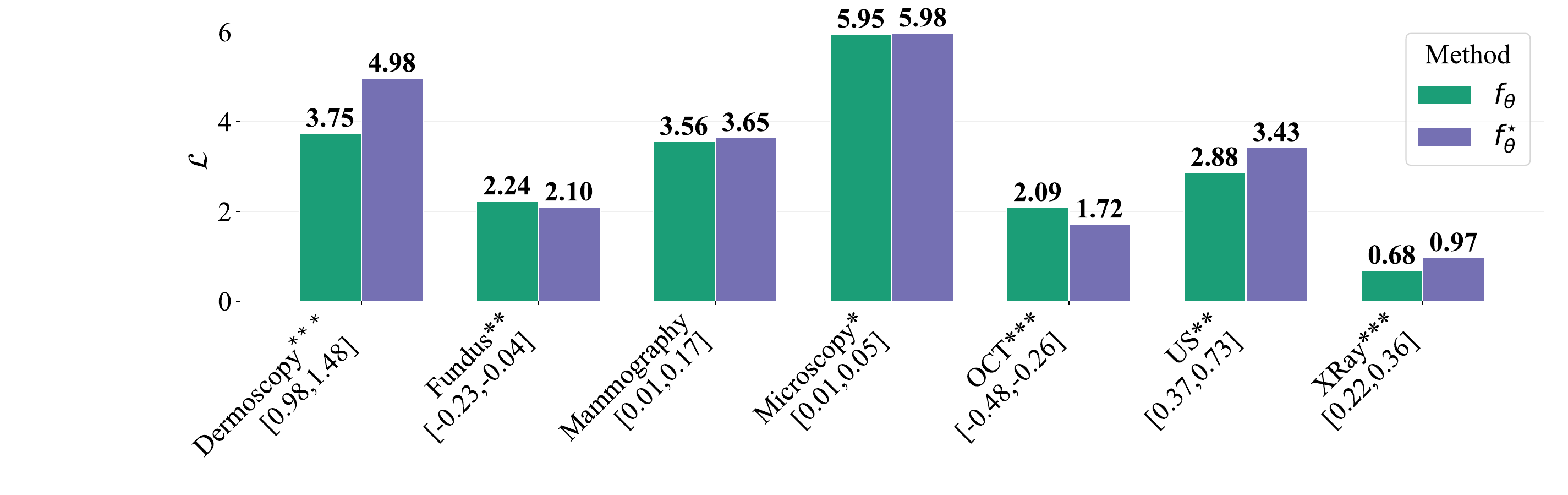}
    \caption{Loss $\mathcal{L}=\mathcal{L}_{BCE}+\mathcal{L}_{Dice}+\mathcal{L}_{IoU}$ for pipelines using raw or edge-enhanced data in pre-training across the medical modalities. 95\% confidence intervals for the mean difference are given.\\
        {\tiny
        * p<0.05, ** p<0.01, *** p<0.001}}
    \label{fig:losses}
\end{figure}

\begin{figure}[b]
        \centering
        \includegraphics[width=1\linewidth]{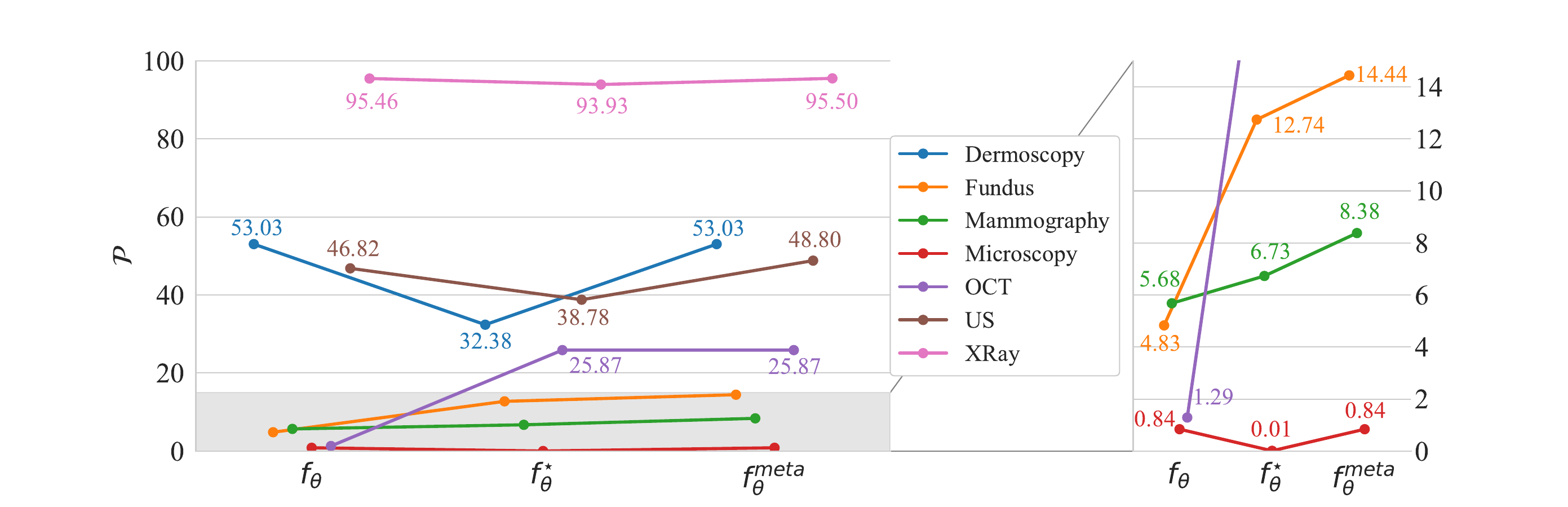}
        \caption{Performance $\mathcal{P}$ across the modalities and the pipelines $f_{\theta}$, $f_{\theta}^{\star}$, and $f^{meta}_{\theta}$.}
        \label{fig:results}
\end{figure}

In this section, we present results of our comprehensive analysis on differences between $f_{\theta}$ and $f_{\theta}^{\star}$, as well as our meta-level approach $f^{meta}_{\theta}$. 
\begin{table}[h]
\caption{Regression coefficients between meta-features and $\Delta\mathcal{P}$. Exemplarily, an increase in entropy of 1 increases $\Delta\mathcal{P}$ for OCT by $15.33$.\\
        {\tiny
        * p<0.05, ** p<0.01, *** p<0.001}}
        \centering
        \begin{tabular}{r|lll}
        \textbf{Modality}
            & \textbf{$r_{\sigma}$} & \textbf{$r_{\varepsilon}$}  \\
        \hline
        Dermoscopy  & $-0.19$ & $-5.26$ \\
        Fundus  &  $-0.17$ & $+2.30$  \\
        Mammography & $+0.12$ &  $+4.22$ \\
        Microscopy & $+0.01$  & $-0.30$\\
        OCT & $+0.04$ &  $+15.33^{\star\star\star}$\\
        US  & $+0.31$ &  $+8.73^{\star}$\\
        XRay &  $+0.04$ &  $-1.96^{\star}$ \\
        \end{tabular}
        
        \label{tab:results}
\end{table}
In~\cref{fig:losses}, we give aggregated loss values $\mathcal{L}$ as well as significant differences in mean using a relative t-test.
Differences of mean exhibited significance for the modalities Dermoscopy, Fundus, Microscopy, OCT, US, and XRay. Further, we provide regression coefficients of the meta-features $\sigma$ and $\varepsilon$ with the absolute performance difference $\Delta\mathcal{P}=\mathcal{P}(f_{\theta}^{\star})-\mathcal{P}(f_{\theta})$ in~\cref{tab:results}. Using these regression coefficients, we can identify trends towards the superior segmentation pipeline based on the meta-features. For standard deviation, no significant correlations are found. However, OCT, US, and XRay data exhibits significant correlation between $\varepsilon$ and $\Delta\mathcal{P}$. While these observed correlations provide valuable insights, the overall pattern does not fully align with our initial predictions. Each hypothesis finds partial support depending on the specific imaging modality. The observed fluctuations in correlation indicate that no single meta-feature can universally predict whether edge-enhanced pre-training improves segmentation quality. These findings confirm the value of a modality-wise meta-classification pipeline that adaptively selects the most appropriate pre-training strategy.
Implementing this meta-classifier, we display the performance $\mathcal{P}$ across all examined modalities in~\cref{fig:results}. 
Across all modalities, the most discriminative meta-feature is the entropy due to the strongest correlations between $\varepsilon$ and the performance difference $\Delta\mathcal{P}$.
For ease of comparison, we give absolute and relative performance differences in~\cref{tab:perfmetamodel}. Additional to $\Delta\mathcal{P}$, we give the relative error reduction $\Delta_{rel}\mathcal{P}=\frac{\mathcal{P}(f_{\theta}^{\star})-\mathcal{P}(f_{\theta})}{\mathcal{P}(f_{\theta})}$. For comparison of the meta-model, we give the absolute performance difference $\Delta\mathcal{P}^{meta}=\mathcal{P}(f^{meta}_{\theta})-\max(\mathcal{P}(f_{\theta}^{\star}),\mathcal{P}(f_{\theta}))$ between the meta-model and the superior pipeline of the one pre-trained on raw or on edge-enhanced data. Further, the relative error reduction $\Delta_{rel}\mathcal{P}^{meta}=\frac{\Delta\mathcal{P}^{meta}}{\max(\mathcal{P}(f_{\theta}^{\star}),\mathcal{P}(f_{\theta}))}$ is presented. 
Across the modalities Fundus, Mammography, US, and XRay, performance can be improved by incorporating the modality-wise meta-classification model. However, for Dermoscopy, Microscopy, and OCT data, we cannot achieve performance gains. 
\begin{table}[t]
\caption{Performance, as well as absolute and relative differences. Maximum score is 100. "Aggregated" is weighted by number of images per modality.}
    \centering
    \fontsize{8}{9}\selectfont
    \begin{tabular}{r|c|c|c|c|c|c|c}
        \textbf{Modality} & $\mathcal{P}(f_{\theta})$ & $\mathcal{P}(f_{\theta}^{\star})$ & $\Delta\mathcal{P}$ & $\Delta_{rel}\mathcal{P}$ & $\mathcal{P}(f^{meta}_{\theta})$ & $\Delta\mathcal{P}^{meta}$ & $\Delta_{rel}\mathcal{P}^{meta}$ \\
        \hline
         Dermoscopy & \textbf{53.03} & 32.38 & -20.65 & -38.94\% & 53.03 & 0.00 & 0.00\%  \\
         Fundus & 4.83 & 12.74 & +7.91 & +163.77\% & \textbf{14.44} & +1.70 & +13.34\% \\
         Mammography & 5.68 & 6.73 & +1.05 & +18.49\% & \textbf{8.38} & +1.65  & +24.52\%  \\
         Microscopy & \textbf{0.84} & 0.01 & -0.83 & -98.81\% & 0.84 & 0.00 &  0.00\%  \\
         OCT & 1.29 & \textbf{25.87} & +24.58 & +1905\% & 25.87 & 0.00 & 0.00\%  \\
         US & 46.82 & 38.78 & -8.04 & -17.17\% & \textbf{48.80} & +1.98 &  +4.23\%  \\
         XRay & 95.46 & 93.93 & -1.53  & -1.60\% & \textbf{95.50} & +0.04 &  +0.04\% \\
         \hline
         Aggregated & 29.59 & 30.32 & +0.8 & +2.70\% & \textbf{35.30} & +4.98 &  +16.42\% \\
    \end{tabular}
    
    \label{tab:perfmetamodel}
\end{table}

\section{Discussion \& Conclusion}
\begin{figure}[b]
    \centering
    \includegraphics[width=1\linewidth]{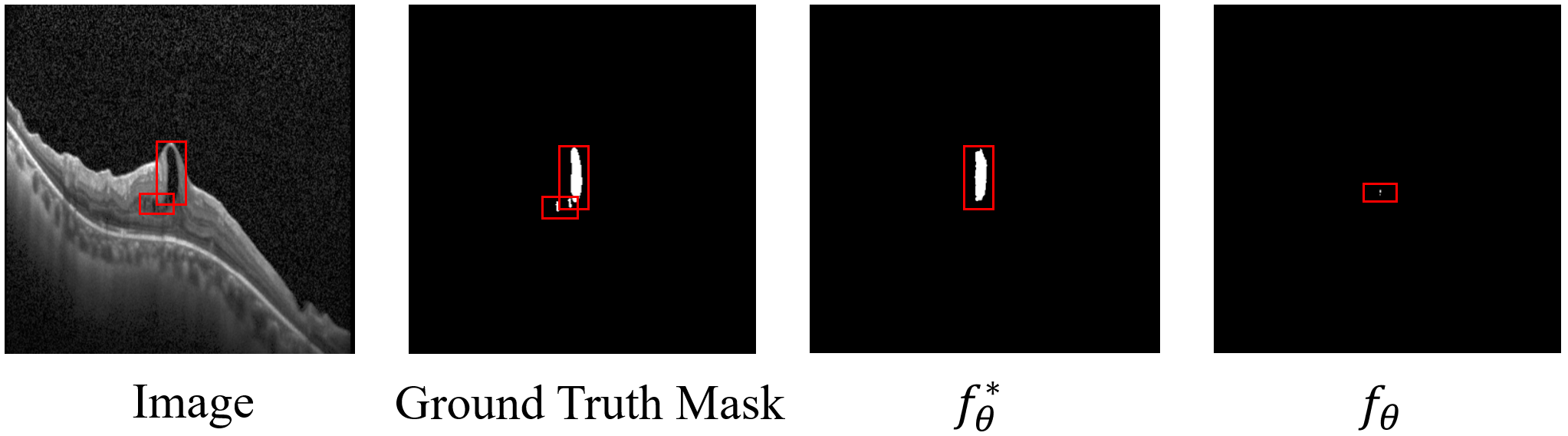}
    \caption{Segmentation masks of OCT data, where $f_{\theta}^{\star}$ outperforms $f_{\theta}$.}
    \label{fig:kirsch_superior}
\end{figure}
\begin{figure}[t]
    \centering
    \includegraphics[width=1\linewidth]{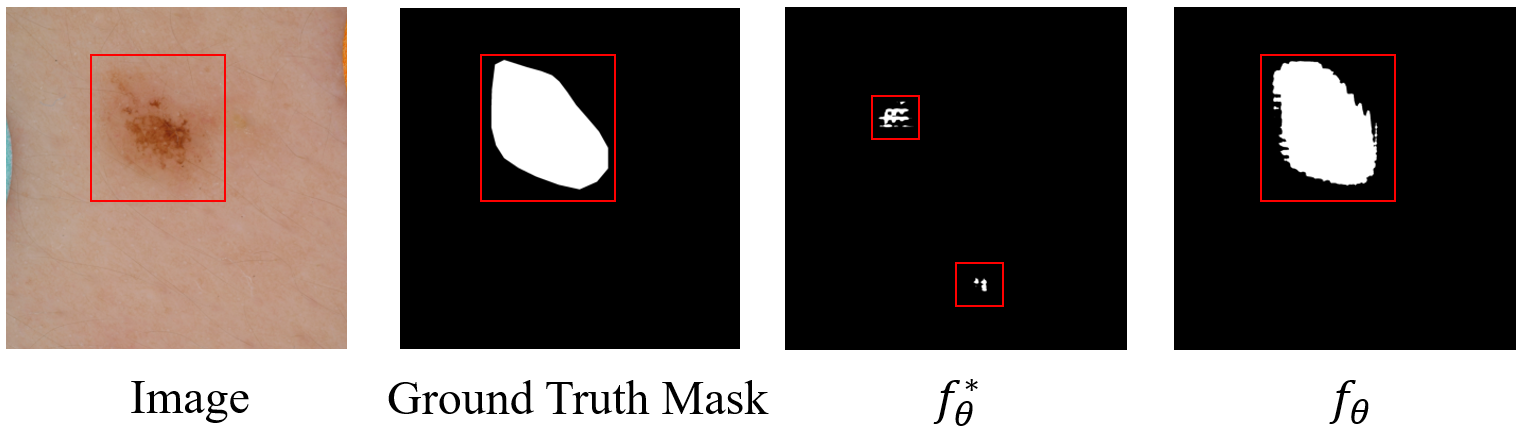}
    \caption{Segmentation masks of Dermoscopy data, where $f_{\theta}$ outperforms $f_{\theta}^{\star}$.}
    \label{fig:raw_superior}
\end{figure}
The results show that edge-enhanced pre-training yields significant performance improvements in certain imaging modalities while degrading segmentation quality in others. Exemplary cases, where both $f_{\theta}^{\star}$ and $f_{\theta}$ significantly outperform each other can be seen in~\cref{fig:kirsch_superior} and~\cref{fig:raw_superior}. This variability highlights the importance of aligning preprocessing strategies with the inherent characteristics of each modality rather than applying a uniform approach. Through the analysis of statistical meta-features---standard deviation and image entropy---, we identified indicators for anticipating when edge emphasis would be advantageous. 
Leveraging these indicators, we developed a selection mechanism that outperforms generic pre-training strategies, optimizing segmentation accuracy. 
While we focussed on the meta-features standard deviation and entropy due to the closeness to pixel intensities, which are determined by the use of edge operators, we encourage further work on higher-dimensional methods for meta-classification like a shallow convolutional neural network or an image encoder. 
Given that no universally optimal edge filter exists for all images~\cite{marr1980theory}, achieving consistent cross-modality improvements in segmentation quality presents a considerable challenge.
With the specific edge kernel used in data pre-processing for foundation models vastly determining performance gains and losses, different edge kernels like Sobel~\cite{kanopoulos1988design} or Prewitt~\cite{prewitt1970object} should be examined to identify influences in performance. 
Since applying edge enhancement in pre-training of foundation models exhibiting great performance increase, we suggest further research on the application of edge enhancement in the subsequent fine-tuning of foundation models.
Integrating multiple filters within a unified model selection framework could also improve adaptability. Alignment between certain edge filters and certain imaging modalities pose vast potential for futher increase in segmentation performance. 
Moreover, applying edge enhancement not only at the pre-training stage but also during subsequent refinements remains an avenue for future work. 
While we validated the benefits and restrictions of using edge-enhanced data in pre-training of foundation models, we recommend further work on increasing computation efficiency due to increased training needs for two foundation models. 
In this paper, we propose applying edge enhancement on image data in pre-training of foundation models for improved cross-modality segmentation quality. 
By leveraging meta-features such as standard deviation and image entropy, we develop a meta-classification strategy to selectively apply edge-enhanced pre-training, achieving an cross-modality segmentation performance improvement by 16.42\% compared to models pre-trained on edge-enhanced data only and 19.30\% compared to models pre-trained on raw data only.
For the modalities Fundus, Mammography, US, and XRay, vast performance gains can be achieved using our meta-classification strategy, while for OCT we recommend solely using edge-enhanced data in pre-training. 
Our work provides a principled framework for optimizing segmentation performance by aligning pre-training strategies with the intrinsic properties of medical imaging modalities, offering a significant step forward in the development of robust, generalizable foundation models for medical image analysis.

\section{Code Availability}
The full source code is available at https://github.com/PaulZaha/CMMC.

\bibliographystyle{splncs04}
\bibliography{references}

\end{document}